\documentclass[11pt,a4paper]{article}

\usepackage[hyperref]{naaclhlt2018}
\usepackage{times}
\usepackage{latexsym}

\usepackage{url}

\usepackage{booktabs}

\usepackage{graphicx}

\usepackage{multirow}

\usepackage{amsfonts}

\usepackage[T1]{fontenc}

\usepackage{microtype}

\aclfinalcopy 

\newcommand{\ex}[1]{\textit{#1}}
\newcommand{\forfinal}[1]{} 

\title{Evaluating historical text normalization systems: \\How well do they generalize?}

\author{Alexander Robertson \\
  School of Informatics \\
  University of Edinburgh \\
  {\tt alexander.robertson@ed.ac.uk} \\\And
  Sharon Goldwater \\
  School of Informatics \\
  University of Edinburgh \\
  {\tt sgwater@inf.ed.ac.uk} \\}

\date{}

\begin{document}
\maketitle
\begin{abstract}

  We highlight several issues in the evaluation of historical text normalization systems that make it hard to tell how well these systems would actually work in practice---i.e., for new datasets or languages; in comparison to more naïve systems; or as a preprocessing step for downstream NLP tools. We illustrate these issues and exemplify our proposed evaluation practices by comparing two neural models against a naïve baseline system. We show that the neural models generalize well to unseen words in tests on five languages; nevertheless, they provide no clear benefit over the naïve baseline for downstream POS tagging of an English historical collection. We conclude that future work should include more rigorous evaluation, including both intrinsic and extrinsic measures where possible.

\end{abstract}

\section{Introduction}\label{intro}

Historical text normalization systems aim to convert 
historical wordforms to their modern equivalents, in order to make historical documents more searchable or to improve the performance of downstream NLP tools.
In historical texts, a single word type may be realized with several different orthographic forms, which may not correspond to the modern form. For example, the modern English word \ex{said} might be realized as \ex{sayed}, \ex{seyd}, \ex{said}, \ex{sayd}, etc.
Spellings change over time, but also vary within a single time period and even within a single author, since orthography only became standardized in many languages fairly recently.

Over the years, researchers have proposed normalization methods based
on rules and/or edit distances \cite{baron2008vard2,bollmann2012automatic,hauser2007unsupervised,bollmann2011applying,pettersson2013normalisation,Mitankin:2014,pettersson2014multilingual}, statistical machine translation \cite{pettersson2013smt,scherrer-erjavec:2013:BSNLP}, and most recently neural network models \cite{bollmann2016,bollmann2017,korchagina2017normalizing}.
However, most of these systems have been developed and tested on a single language (or even a single corpus), and
many have not been compared to the naïve but strong baseline that only changes words seen in the training data, normalizing each to its most frequent modern form observed during training.\footnote{Some authors have focussed on {\em unsupervised} normalization, where the naïve baseline is to leave words unchanged \cite{Mitankin:2014,hauser2007unsupervised}. We consider only {\em supervised} systems in the remainder of this paper.} These issues make it hard to tell which methods generalize across languages and corpora, and how they compare to each other. Moreover, researchers have rarely examined whether their systems actually improve performance on downstream tasks.

This paper brings together best practices for evaluating historical text normalization systems, highlighting in particular the need to report results on unseen tokens and to consider the naïve baseline. We focus our evaluation on two recent neural models: 
one that has been previously tested only on a German collection that is not widely available \cite{bollmann2017}, and one that is adapted from work on morphological re-inflection, but has not been used for historical text normalization \cite{hardattention}. Both are encoder-decoder models; the former with soft attention, and the latter with hard monotonic attention.
\forfinal{We use both models out of the box, with no hyperparameter retuning.}

We present results on five languages,
for both seen and unseen words and for various amounts of training data.
The soft attention model performs surprisingly poorly on seen words, so that its overall performance is worse than the naïve baseline and several earlier models \cite{pettersson2014multilingual}. However, on unseen words (which we argue are what matters), both neural models do well.

Unfortunately, these positive results did not translate into improvements when we tested the English-trained models on a downstream POS tagging task using a different historical collection spanning a similar time range. Normalizing the text gave better tag accuracy than not normalizing, but neither neural model convincingly outperformed the naïve normalizer.
Although these results are disappointing, the clear evaluation standards laid out here should benefit future work in this area.

\section{Task setting and issues of evaluation}

We follow previous work in training our systems on pairs $(h,m)$ of historical tokens and their gold standard modern forms.\footnote{It would be possible to train on full texts rather than isolated tokens, which could improve results for ambiguous forms. However, previous models have not addressed this setting, nor do we, leaving this for future work.}
Note that at test time, most of the $h$ tokens will have been seen before in the training data (due to Zipf's law), and for these tokens it is very difficult to beat a baseline that normalizes each $h$ to the most common $m$ seen for it in training.\footnote{Our version breaks ties by choosing the first $m$ observed.}
Thus, in practice, normalization systems should typically only be applied to {\em unseen} tokens. It is therefore critical to report
both dataset statistics and experimental results for unseen tokens.

Unfortunately, some recent papers have only reported accuracy on all tokens, and only in comparison to other (non-baseline) systems \cite{bollmann2016,bollmann2017,korchagina2017normalizing}. These figures can be misleading if systems underperform the naïve baseline on seen tokens (which we show does happen in practice). To see why, suppose 80\% of test tokens were seen in training, and the baseline gets 90\% of them right, while system A gets 80\% and system B gets only 70\%. Meanwhile the baseline gets only 50\% of unseen tokens right, whereas systems A and B get 70\% and 90\%, respectively. A's accuracy is higher {\em overall} than B's (78\% vs 74\%), but {\em both} systems underperform the baseline (82\%).
More importantly, the best system (90\% accuracy overall) is achieved by applying the baseline to seen tokens, and the system that generalizes best (B) to unseen tokens;
it is irrelevant that A scores higher overall than B.

Stemming from the reasoning above, we argue that a full evaluation of any spelling normalization system requires more complete dataset statistics and experimental results.
In describing the training and test sets, researchers should not only report the number of {\bf types} and {\bf tokens}, but also the percentage of {\bf unseen tokens} in the test (or dev) set and the percentage of {\bf training items $(h,m)$ where $h=m$}. This last statistic measures the degree of spelling variation, which varies considerably between corpora. 

As for reporting results, we have argued that accuracy should be reported separately for {\bf seen vs unseen tokens}, and overall results compared to the {\bf naïve memorization baseline}.
Since historical spelling normalization is typically a low-resource task, systems should also ideally be tested with {\bf varying amounts of training data} to assess how much annotation might be required for a new corpus \cite{pettersson2014multilingual,bollmann2016,korchagina2017normalizing}. Finally, since these systems may be deployed on corpora other than those they were trained on, and used as preprocessing for other tasks, we advocate reporting {\bf performance on a downstream task and/or different corpus}. To our knowledge the only previous supervised learning system to do so is \citet{pettersson2013smt}.

\section{Models}

We focus on two neural encoder-decoder models for spelling normalization, comparing them against the memorization baseline and to previous results from \citet{pettersson2014multilingual}.
The first model \cite{bollmann2017}\footnote{https://bitbucket.org/mbollmann/acl2017} uses a fairly standard architecture with a bi-directional LSTM encoder and an LSTM decoder with soft attention \cite{xu2015show}, and is trained using cross-entropy loss.

The second model is a new approach to spelling normalization, which adapts the morphological reinflection system of \newcite{hardattention}.\footnote{https://github.com/roeeaharoni/morphological-reinflection} The reinflection model generates the characters in an inflected wordform $(y_{1:n})$, given the characters of its lemma $(x_{1:m})$ and a set of corresponding morphological features $(f)$. 
Rather than using a soft attention mechanism that computes a weight vector over the entire sequence, this model
exploits the generally monotonic character alignment between $x_{1:m}$ and $y_{1:n}$ and attends to only a single encoded input character at a time during decoding. 

Architecturally, the model uses a standard bi-directional encoder. The decoder steps through the characters of the input and considers jointly the output of the previous step, the morphological features, and the currently attended encoded input. It outputs either a character or an advance symbol (to advance the focus of attention for the next time step). It is trained on an oracle sequence of write/advance actions $s_{1:q}$ which are generated from an automatic alignment of the input and output sequences. The model maximizes $p(s_{1:q} | x_{1:m}, f)$. For details, see \newcite{hardattention}.

We adapt the model to our purpose by removing the morphological features $f$, maximising only $p(s_{1:q} | x_{1:m})$.
The monotonic assumption is well-suited to our task, since
fewer than 0.4\% of edit operations require non-monotonic alignments (i.e. character transpositions) in
any of our datasets.

Other than removing the need for morphological features from the hard attention model, and increasing the number of training epochs to 50 for both models, we did no further hyperparameter tuning, since our goal was to assess the ``off-the-shelf" performance of these systems.

\section{Experiments}

We use the same datasets as \newcite{pettersson2014multilingual}, with data from five languages over a range of historical periods.\footnote{English: \newcite{markus1999manual}; German: \newcite{scheible2011gold};  Hungarian: \newcite{simon2014corpus}; Icelandic: \newcite{rognvaldsson2012icelandic}; Swedish: \newcite{fiebranz2011making}. For details of their dates and contents, see \newcite{pettersson2014multilingual}.} We use the same train/dev/test splits as Pettersson; dataset statistics are shown in Table \ref{description}. Because we do no hyperparameter tuning, we do not use the development sets, and all results are reported on the test sets.

\begin{table}[]
\resizebox{\columnwidth}{!}{%
\begin{tabular}{@{}lrrrrr@{}}
  \toprule
 & Tokens & $h$ typ & $m$ typ & \%nc & \%uns \\ \midrule
Eng & 148/16/17k & 19.4k & 10.6k & 73.9 & 8.6 \\
Ger & 39/5/5k & 9.0k & 8.4k & 84.8 & 14.8 \\
Hun& 137/17/17k & 45.5k & 25.8k & 15.4 & 24.1 \\
Ice & 52/6/6k & 9.7k & 8.5k & 48.0 & 11.3 \\
Swe & 28/2/34k & 8.3k & 6.5k & 65.9 & 22.4 \\ \bottomrule
\end{tabular}%
}
\caption{Dataset statistics: the number of tokens in train/dev/test sets; {\it h}istorical and {\it m}odern word types and \% of ``no-change'' tokens ($h=m$) in the training sets; and the \% of dev set tokens that are unseen in training.}
\label{description}
\end{table}

Each system was tested as recommended above, with accuracy reported separately on seen and unseen items, and for different training data sizes.
To evaluate the downstream effects of normalization, we applied the models to a collection of unseen documents and then tagged them with the Stanford POS tagger, which comes pre-trained on modern English.
The documents are from the Parsed Corpus of Early English Correspondence (PCEEC) \cite{taylor2006york}, comprised of 84 letter collections from the 15th-17th centuries. (Our English normalization training data is from the 14th-17th centuries.) PCEEC contains roughly 2.2m manually POS-tagged tokens but no spelling annotation. Because it uses a large and somewhat idiosyncratic set of POS tags, we converted these to better match the Stanford tags before evaluating (though the match still isn't perfect; accuracy would be higher in all cases if the tag sets were identical). Baselines are provided by tagging the unnormalized text and the output of the naïve normalization baseline.

\begin{table*}[ht!]
\centering
\resizebox{\textwidth}{!}{%
\begin{tabular}{@{}lccc|ccc|ccc|ccc|ccc@{}}
\hline
                     & \multicolumn{3}{c|}{English}                                                       & \multicolumn{3}{c|}{German}                                    & \multicolumn{3}{c|}{Hungarian}                                 & \multicolumn{3}{c|}{Icelandic}                                 & \multicolumn{3}{c}{Swedish}                                   \\ \hline
\multicolumn{1}{c}{} & \multicolumn{1}{c}{A} & \multicolumn{1}{c}{S} & \multicolumn{1}{c|}{U} & \multicolumn{1}{c}{A} & \multicolumn{1}{c}{S} & \multicolumn{1}{c|}{U} & \multicolumn{1}{c}{A} & \multicolumn{1}{c}{S} & \multicolumn{1}{c|}{U} & \multicolumn{1}{c}{A} & \multicolumn{1}{c}{S} & \multicolumn{1}{c|}{U} & \multicolumn{1}{c}{A} & \multicolumn{1}{c}{S} & \multicolumn{1}{c}{U} \\ \hline
Hybrid 		     & 92.9                      &                          &                            & 95.1  &                          &                            & 76.4  &                          &                            & {\bf 84.6}  &                    &                            & 90.8  &                          &                            \\
GIZA++ un            & {\bf 94.3}                &                          &                            & {\bf 96.6}  &                    &                            & 79.9  &                          &                            & 71.8  &                          &                            & {\bf 92.9}  &                    &                            \\
GIZA++ bi            & 92.4                      &                          &                            & 95.5  &                          &                            & 80.1  &                          &                            & 71.5  &                          &                            & 92.5  &                          &                            \\ \hline
Mem.\ baseline       & 91.5                      & {\bf 96.9}               & 30.5                       & 94.1  & 96.9                     & 30.5                       & 73.6  & {\bf 96.0}               & 2.9                        & 80.3  & {\bf 86.8}               & 28.3                       & 85.4  & {\bf 98.1}               & 41.4                       \\
Soft attention       & 89.9                      & 93.7                     & 46.9                       & 94.3  & 98.1                     & 72.4                       & 79.8  & 89.4                     & 49.6                       & 83.1  & 85.9                     & 60.1                       & 89.7  & 97.2                     & 63.8                       \\
Hard attention       & 93.0                      & 96.6                     & {\bf 52.4}                 & 96.5  & {\bf 99.3}               & {\bf 80.5}                 & {\bf 88.0}  & 95.3               & {\bf 65.0}                 & 83.5  & 86.2                     & {\bf 61.4}                 & 90.7  & 97.9                     & {\bf 65.7}                       \\ \hline
\end{tabular}%
}
\caption{Tokens normalized correctly (\%) for each dataset. Upper half: results on (A)ll tokens reported by \citet{pettersson2014multilingual} for a hybrid model (apply memorization baseline to seen tokens and an edit-distance-based model  to unseen tokens) and two SMT models (which align character unigrams and bigrams, respectively). Lower half: results from our experiments, including accuracy reported separately on (S)een and (U)nseen tokens.
}
\label{comparison-results}
\end{table*}

\begin{figure*}[]
\centering
\resizebox{0.99\textwidth}{!}{%
\includegraphics{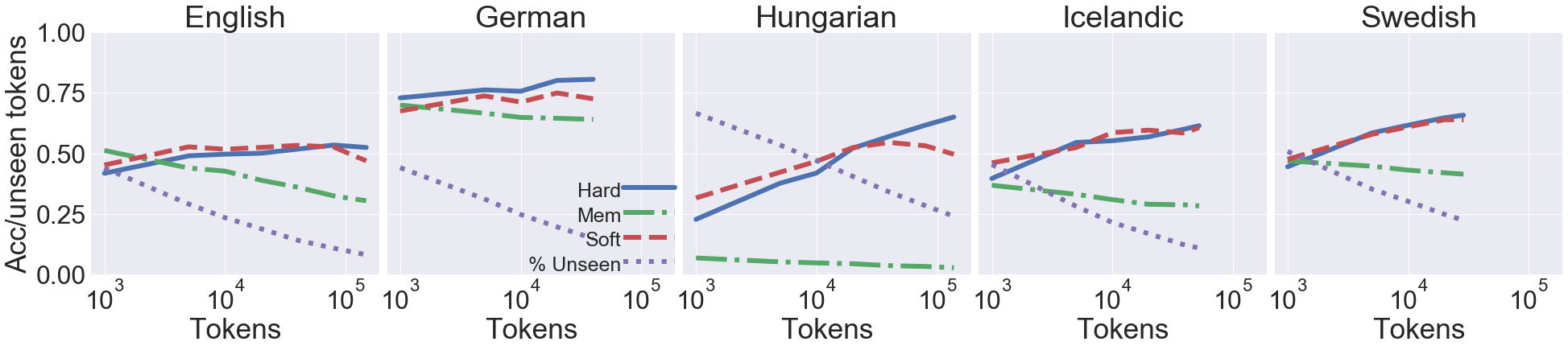}
}
\caption{Proportion of unseen tokens, and normalization accuracy on those tokens, as training data size is varied.
}
\label{fig:unseen_accuracy}
\end{figure*}

\paragraph{Results: normalization accuracy}

Table \ref{comparison-results} gives test set results for all models, broken down into seen and unseen items where possible.
\footnote{We obtained our datasets from \citeauthor{pettersson2014multilingual} but our baseline results are slightly different from what they report. The differences (theirs--ours) are -0.1, 0.2, 0.4, 1.2, 0.6 for Eng, Ger, Hun, Ice, Swe respectively. This could be due to differences in tie-breaking methods, or to another unknown factor. These differences suggest using caution in directly comparing their non-baseline results to ours.}
 The split into seen/unseen highlights the fact that neither of the neural models does as well on seen items as the baseline; indeed the soft attention model is considerably worse in English and Hungarian, the two largest datasets.\footnote{When we varied the training data sizes, we found that the soft attention model actually gets {\em worse} on seen tokens in all languages as the training data increases beyond a relatively small size. We have no good explanation for this, and it's possible that tuning the parameters would help.} The result is that this model actually underperforms the baseline when applied to all tokens, although a hybrid model (baseline for seen, soft attention for unseen) would outperform the baseline.
Nevertheless, the hard attention model performs best on unseen tokens in all cases, often by a wide margin, and also yields competitive overall performance.

We also compared the accuracy of the two neural models at different training data sizes starting from 1k tokens. On {\em seen} tokens, the baseline was best in all cases except for 1k tokens in Hungarian and Icelandic (where the soft attention model was slightly better) and the largest two data sizes in German (where the hard attention model was slightly better). This supports our claim that
learned models should typically only be applied
to {\em unseen} tokens.

Accuracy on unseen tokens is shown in Figure \ref{fig:unseen_accuracy}.
Note that the set of unseen items gets smaller and presumably more difficult as training data size increases, so the baseline gets worse. In contrast, the neural models are able to maintain or increase performance on this set. We expected that the bias toward monotonic alignments would help the hard attention model at smaller data sizes, but it is the soft attention model that seems to do better there, while the hard attention model does better in most cases at the larger data sizes. 
Note that \citet{bollmann2017} trained their model on individual manuscripts, with no training set containing more than 13.2k tokens. The fact that this model struggles with larger data sizes, especially for seen tokens, suggests that the default hyperparameters may be tuned to work well with small training sets at the cost of underfitting the larger datasets.

\begin{figure}[]
\centering
\includegraphics[width=.9\columnwidth]{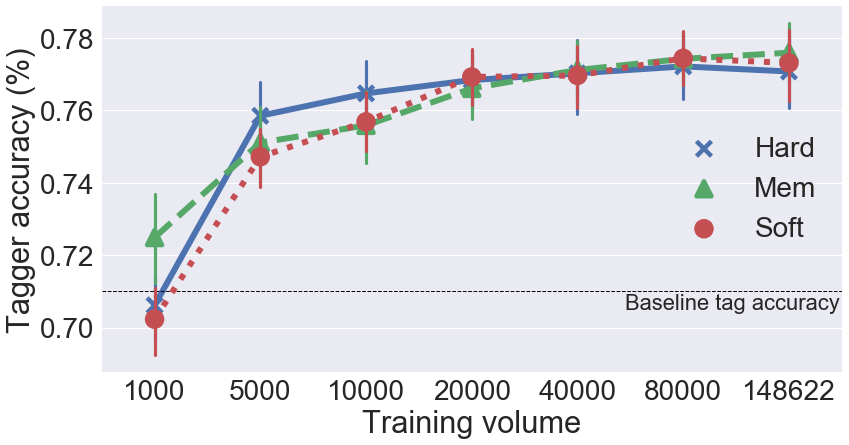}
  \caption{Average POS tagging accuracy on the unnormalized PCEEC texts (bottom of plot) and using three different normalization methods, as a function of the amount of data used to train the normalization systems.
  }
  \label{posresults}
\end{figure}

\paragraph{Results: POS tagging}

Based on our results above, we tested the neural models by applying them only to unseen tokens in the PCEEC, and normalizing seen tokens using the naïve baseline in all cases. The PCEEC is a heterogeneous collection, so baseline tagger accuracy on the unnormalized text ranges from 52.0\% to 82.6\%, with an average of 71.0\% ($\sigma$: 6.8). Figure \ref{posresults} shows the effects of normalizing using the different methods.

Although normalizing provides a clear benefit, in most cases the neural models are no better than normalizing using the baseline method. The exception is at 5k and 10k training items, where a two-tailed t-test shows that the hard attention model is significantly better than the other methods ($p<0.01$). We also tried preprocessing both the normalization and tagging datasets by lowercasing all tokens; this resulted in small improvements in most cases (about 1 point) but any remaining differences were to the benefit of the baseline method.

Our findings differ from those of \citet{pettersson2013smt}, who reported that their SMT-based system did work better than the baseline normalizer for POS tagging in Icelandic and verb identification in Swedish. Our contrasting findings could derive either from our use of different models or different datasets; nevertheless, they highlight the fact that intrinsic improvements do not always translate into extrinsic ones.

\section{Conclusion}

We have highlighted some important issues in the evaluation of historical text normalization systems: in particular, the need to report accuracy on unseen tokens and to compare performance to a naïve memorization baseline. Following these recommendations, we evaluated two neural models, one of which is new to this task. Across five languages, both models greatly outperformed the baseline on unseen tokens, with the soft attention model doing a bit better for smaller data sizes, and the hard attention model doing a bit better for larger ones. However, these improvements did not translate into clearly better POS tagging downstream.

Despite these mixed results, we hope that the evaluation guidelines presented here will help promote work in this area, in order to eventually provide better tools for working with historical text collections.

\section{Acknowledgements}

We thank \citeauthor*{pettersson2014multilingual} for the datasets, and \citeauthor*{hardattention} and \citeauthor*{bollmann2017} for making their code available.
This work was supported in part by the EPSRC Centre for Doctoral Training in Data Science, funded by the UK Engineering and Physical Sciences Research Council (grant EP/L016427/1) and the University of Edinburgh.

\bibliography{paper_nocomments}
\bibliographystyle{acl_natbib}

\end{document}